# Generative Adversarial Networks for Dental Patient Identity Protection in Orthodontic Educational Imaging

**Mingchuan TIAN[1], Wilson Weixun LU[2], Kelvin Weng Chiong FOONG[3], Eugene LOH[4]**
**E-mail:** mingchuantian1@gmail.com[1], wilson_weixun_lu@nuhs.edu.sg[2], denfwc@nus.edu.sg[3], denlohe@nus.edu.sg[4]
Correspondence should be addressed to mingchuantian1@gmail.com
**Address:** Faculty of Dentistry, National University of Singapore, National University Centre For Oral Health Singapore, 9 Lower Kent Ridge Rd, Singapore 119085

## Abstract

**Objectives:** Dental patient images are critical educational resources for dental schools and research institutions. However, privacy concerns necessitate the occlusion of identifiable features, primarily around the eye area, prior to these images being used for instructional or research purposes. This research aimed to present an innovative area-preserving Generative Adversarial Networks (GAN) inversion approach, capable of preserving dental features while altering facial identities in dental patient images, with the dual purpose of protecting patient privacy and providing valuable resources for dental education and research.
**Methods:** This study first addressed enhancements to the existing GAN Inversion technique to optimize the preservation of dental features in the synthesized images. Subsequently, an encompassing technical framework was constructed, incorporating several deep learning models, to serve as end-to-end development guidance and a practical image identity swap application.
**Results:** The system was tested with a diverse range of inputs, including smiling and non-smiling frontal facial pictures, commonly used in diagnosing skeletal asymmetry and facial anomalies. Results demonstrated that our approach successfully adapted the context from one image to another, keeping the results seamlessly compatible, while preserving dental features such as teeth, jaw shape, and lips for subsequent oral diagnosis and dental education. The effectiveness of our approach was further validated through an evaluation by a panel of five experienced clinicians who conducted an analysis on 30 original images, as well as the 30 GAN-processed images. The images generated by our application were observed to achieve effective de-identification while simultaneously preserving the realism of essential dental features. Additionally, these images were assessed to be of moderate utility for dental diagnostics and education, underlining the practical applicability of our technique in real-world scenarios.
**Conclusions:** The proposed area-preserving GAN inversion approach and the end-to-end framework showed promising results in preserving dental features while modifying facial identities, hence, providing a potential avenue for privacy-protected image utilization in dental education and research.

**Clinical Significance:** The innovative area-preserving GAN model and the comprehensive identity-swapping framework can expedite the de-identification process of dental patient images, reducing labor and improving efficiency in dental education. This method improves the diagnostic acumen of the students by being able to provide them with more exposure to the variations in orthodontic malocclusions. Moreover, this method facilitates the generation of de-identified datasets for further 2D image research at large research institutions.

**Funding:** This work was supported by National University of Singapore, Faculty of Dentistry.

# Introduction

Facial imaging represents a vital resource for education and research, as they offer essential insights into a myriad of dental conditions, including skeletal asymmetry and facial anomalies. However, these frontal facial images inherently carry identifiable characteristics, necessitating careful preservation measures to ensure privacy protections. Traditional methods of de-identification typically involve manual preprocessing, such as affixing strip-shaped occluders over the eye region. However, not only is this process time-consuming and potentially expensive when applied to large datasets, it also fails to provide complete de-identification, as identifiable features such as hair, eyebrows, and ears can still be discernible to the public. Moreover, the manual occlusion process often negatively impacts the overall utility of the images for educational and research purposes, by hindering a comprehensive visual assessment.

In response to this challenge, we propose an innovative, area-preserving Generative Adversarial Network (GAN) inversion approach. This AI-driven methodology aims to safeguard the essential dental features while seamlessly integrating the image with a fresh identity, effectively circumventing the limitations of traditional methods.

Our objective is to significantly enhance the de-identification process of dental patient images, thus creating a more efficient and reliable pipeline for the preparation of these valuable images for both educational and research applications.

# Related work

Generative Adversarial Networks (GANs) represent a distinct class of generative models rooted in deep learning that aim to generate novel data through adversarial training [1]. The architecture of GANs typically involves two distinct neural networks: a generator (G) and a discriminator (D), which undergo joint training within an adversarial framework.

The principal task of the generator is to produce artificial data so compelling that it mimics real data, while the discriminator's function is to accurately distinguish between real and artificial data. The advent of GANs has spurred the development of multiple variants, each striving to augment the quality of synthesized data [2,3,4,5,6] and bolster the stability of training [7,8,9]. In recent years, well-optimized GANs have been found to encapsulate disentangled semantic information within their latent space, thereby facilitating the editing of synthesized images and providing greater control over the generative process. This concept was further developed by StyleGAN [10], which incorporated an additional layer that maps the original Z latent space to the W space, thus setting new benchmarks for image editing performance. Shen et al. [11] made further strides in this field, providing an in-depth interpretation of the semantic face editing potential within the GANs' latent space.

**GAN Inversion Techniques:** Given a trained GAN model, the GAN inversion technique is designed to train an additional encoder with the aim of identifying the most accurate latent code that encapsulates the image within the GAN's latent space. This inverted latent code can subsequently be deployed into the GAN generator to faithfully reconstruct the input image, thus proving instrumental in image editing, as any manipulation of the latent code translates into modifications of the output image. The training process typically employs a pixel-wise reconstruction loss to gauge discrepancies between the input image and the reconstructed output, ensuring that the encoder can aptly map the input into the GAN latent space. A well-adjusted encoder model should therefore be capable of transforming an input image into a latent code, which, when fed into the GAN generator, will accurately reconstruct the original image. Current GAN inversion techniques largely fall into two distinct categories - learning-based and optimization-based. Learning-based techniques initially synthesize a set of images using randomly sampled latent codes, subsequently using these images and codes to train a deterministic model. Conversely, optimization-based techniques focus on individual instances, optimizing the latent code to reduce pixel-wise reconstruction loss.
The progress made in GAN inversion techniques is crucial, given their key role in bridging the real and fake image domains [15, 16, 17, 18, 19, 20].

**Key Distinctions:** A significant shortcoming in current inversion methods is their single-minded focus on the strict preservation of specific areas, while often overlooking the implications for subsequent applications. Existing GAN inversion techniques are largely preoccupied with the semantics learnt by the encoder and the editability of the images, potentially leading to output images that do not closely resemble the input, thereby restricting their practical applicability. Take our dental picture identity-swap application for instance, which demands an accurate representation of dental features, such as teeth, jaw morphology, and lips. In light of this, we contend that relying solely on the latent code-based loss as the evaluation metric during encoder training is inadequate.

To address this, we propose the area-preserving encoder, designed to scrutinize the pixel-wise reconstruction quality of a specific area in order to preserve dental features post-identity merge. Complementing this, we have engineered an end-to-end software pipeline, aiming to deliver greater insight into application development.

# Materials and Methods

## Area-preserving GAN Inversion

Our methodology is designed to preserve the dental area while simultaneously generating a new facial identity from frontal face images, thereby ensuring privacy. This generation process leverages two types of images: the context image and the target image. The context image refers to a frontal face image of a non-existent identity, generated through StyleGAN, while the target image is the original image obtained from dental clinics. To blend these images into a unified output, we first excise the dental area from the target image, subsequently superimposing it onto the corresponding position within the context image prior to the GAN Inversion. This GAN Inversion subsequently maps the merged image into a latent code that holds semantic significance to the GAN generator. Here, 'semantics' denotes the knowledge spontaneously derived by the GAN from the analyzed data [21, 22, 23, 24]. Consequently, the inverted latent code reflects the primary facial features rather than any minor discrepancies between the cropped target dental image and the context identity image. Once this latent code is incorporated into the GAN generator, it is capable of faithfully reconstructing realistically merged results, with both identity and dental features maintained.

## Generator

Prior to the process of training an encoder to invert the image into the GAN latent space, we first outline and train a GAN network. A standard GAN model comprises two elements: a generator G and a discriminator D. The generator $G(\cdot): Z \to X$ synthesizes high-quality images from a random sample z drawn from a standard distribution space Z, while the discriminator $D(\cdot)$ differentiates between real and synthesized data. Throughout training, the generator G is responsible for assigning semantic meanings to the random sampling space Z. This functionality proves beneficial as it enables semantic editing of the synthesized image by simply adjusting the z latent code. It also aids in our subsequent task, as inverting an image into a semantically meaningful latent code contributes to enhanced image reconstruction. For our GAN network, we chose StyleGAN2 [25], predominantly due to its extensive semantics within their latent space. We utilized the widely-adopted CelebA-HQ dataset, which consists of 30,000 high-quality front face images at a resolution of 1024×1024, for initial model training. Given that our subsequent application is specifically focused on dental patient images, we gathered 1,750 real dental patient images from the National University Centre for Oral Health, Singapore (NUCOHS) for further model fine-tuning.

## Area-preserving Encoder

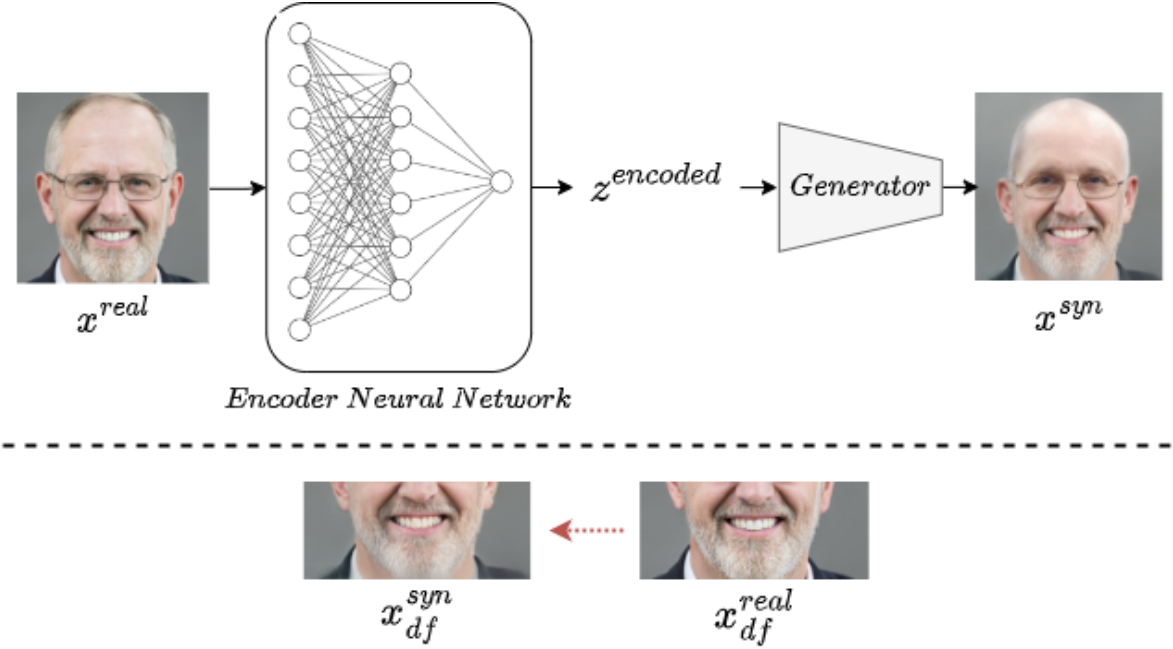

Fig.1. The training process of area-preserving encoders for GAN inversion. The **red** dashed arrow indicates the major supervision

Training an encoder for image-to-code mapping is one of the major challenges in GAN Inversion problems. Existing methods primarily focus on the quality of the encoder-produced latent code, based on which loss functions are imposed. As shown in Equation (1), conventional methods feed a randomly sampled latent code $z_{sam}$ into a trained GAN generator to produce a synthesized image result $x_{syn}$. The image $x_{syn}$ is then fed into the encoder to get $z_{syn}$. The loss function is defined as the euclidean distance between the latent code $z_{sam}$ and the inverted $z_{syn}$. This method allows the encoder to learn a basic mapping from image to GAN's latent space. However, latest research suggests that only comparing the z latent code is ineffective in learning semantic information of the reconstructed image is mostly omitted in the training process. Another problem is that the conventional loss function of the encoder cannot guarantee the exact reconstruction of the dental area.

$$\min_{\Theta_E} \mathcal{L}_E = \|\mathbf{z}^{sam} - E\left(G\left(\mathbf{z}^{sam}\right)\right)\|_2 \qquad (1)$$

To solve the above problems, we propose the Area-preserving Encoder, which focuses more on the reconstruction quality of a particular area. Specifically, we add a loss term that compares the stitched area between the synthesized image and the original image, which enforces the area to be unchanged after generation. This suits our downstream application to preserve dental features after identity generation. To ensure the realisticity of the reconstructed image, we further conduct a pixel-wise comparison between the input and the synthesized image as a whole. Hence, the training process can be formulated as

$$\min_{\Theta_E} \mathcal{L}_E = \|\mathbf{z}^{sam} - E\left(G\left(\mathbf{z}^{sam}\right)\right)\|_2 + \lambda \|\mathbf{x}^{\text{real}} - G\left(E\left(\mathbf{x}^{\text{real}}\right)\right)\|_2 \qquad (2)$$

$$+\lambda_{df} \left\| \mathbf{x}_{df}^{\text{real}} - G\left(E\left(\mathbf{x}_{df}^{\text{real}}\right)\right) \right\|_2$$

where $\theta_E$ denotes the parameters of the encoder. $\lambda_{df}$ and $\lambda_{df}$ are the hyper-parameters for the dental feature loss and the whole image reconstruction loss, respectively. In this way, we penalize the encoder when the resulting image i) does not ensemble the stitched input image, and ii) with a dental area not identical to the patient. The output code is therefore guaranteed to preserve the dental semantic information of the input image.

## Application Framework

Our technical framework is specifically tailored for a downstream identity-swapping application to enhance its utility in dental education. To optimize user experience and generation quality, the comprehensive framework encompasses four core components - a face detector, an identity classifier, a fake ID generator, and an area-preserving image generator. Each component integrates neural networks meticulously designed and trained for unique roles.

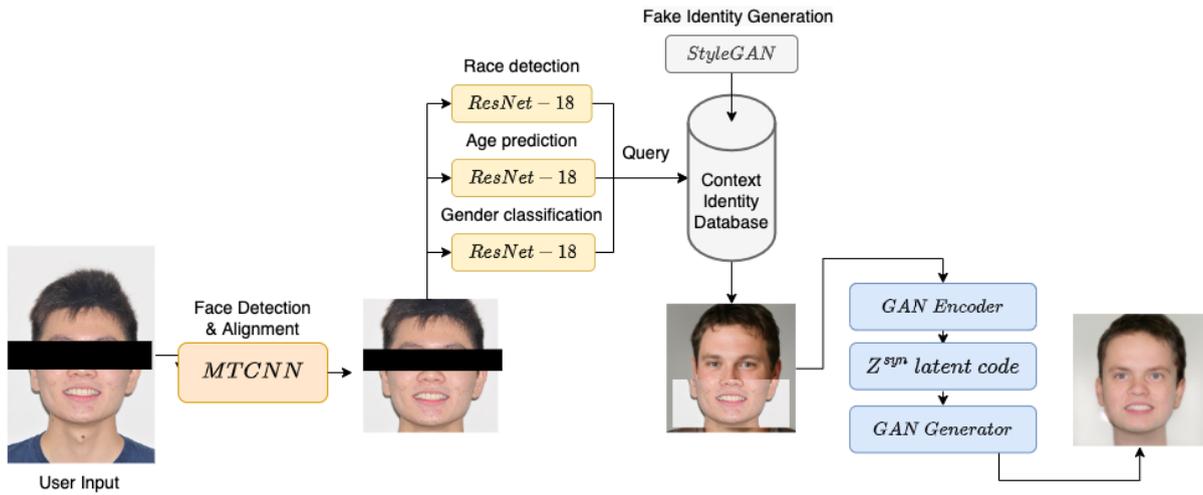

Fig.2. The holistic framework of our proposed solution to face identity-swap application for dental patient privacy protection, consisting of face detection, classification, identity generation, and our Area-preserving GAN inversion. The user input is masked for privacy protection.

## Face Alignment and Detection

Prior to processing and inverting the image into a latent code, we ascertain the presence of a human face and enforce its central alignment. Multi-task Cascaded Convolutional Networks (MTCNN) provide a solution for both face detection and face alignment. A pre-trained MTCNN, when applied to large face datasets, yields face/non-face classification, bounding box location, and facial landmarks. The MTCNN resizes the image to different scales and introduces them into a three-stage cascaded framework to outline each facial element's position. In this study, we incorporated the MTCNN as our pipeline's initial component to guarantee two prerequisites for our Area-preserving Encoder to function as expected: i) the existence of a patient's face in the image ii) standardized positioning of each face within the image.

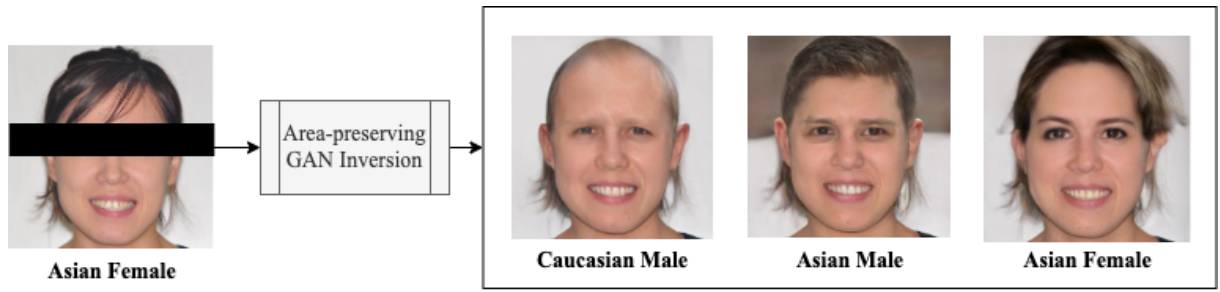

Fig.3. Comparison on area-preserving GAN Inversion with different context images. Area-preserving GAN yields more realistic results when two pictures of similar identities are merged before encoding.

## Identity Information Matching

Our findings indicate that the realism of identity-swapped images can be further enhanced by pairing two images of similar identities. For instance, a dental area image cropped from a young Caucasian male results in a more realistic image when merged with a context image from another Caucasian male of a similar age. Based on these findings, we propose an Identity Information Matching pipeline to pair each dental patient with a related context identity image. We first prepared a Context Image Database with over 300 fake identity images generated by StyleGAN, manually labelling each image with identity properties (Gender, Age, Race). We subsequently trained three separate Convolutional Neural Networks (CNNs) for identity, age, and gender classification. We selected ResNet-18 for our CNN network due to its widespread use in modern classification tasks, allowing us to match the input patient image with the most suitable context identity image for subsequent stitching and synthesis.

# Results

We conducted rigorous evaluations of our area-preserving GAN inversion method in the context of simulated identity protection tasks. To reflect real-world scenarios, synthetic face images created by StyleGAN were used as test inputs for our system. Their high realism and the extensive coverage of key attributes such as race, gender, and facial expressions makes these images ideal proxies for real-world cases. The labels $x^{real}_{female}$ and $x^{real}_{male}$ were assigned to these synthetic images. Images synthesized through our area-preserving GAN inversion approach were assigned the labels $x^{syn}_{female}$ and $x^{syn}_{male}$. For this evaluation, our focus was specifically on images displaying both smiling and non-smiling expressions, as these are most commonly used in the diagnosis of skeletal asymmetry and disproportionate facial structures. To ensure comprehensive assessment of our method's effectiveness across a variety of demographics, we categorized images into four age groups: Child, Teenager, Adult, and Senior.

Our results indicated that the area-preserving GAN inversion method was highly effective at adapting the context from the input image ($x_{female}^{real}$ and $x_{male}^{real}$) to the synthesized image ($x_{female}^{syn}$ and $x_{male}^{syn}$). Additionally, it ensured seamless compatibility between the source and the generated images. Crucially, the preservation of dental features - including teeth, jaw shape, and lips - was successfully achieved. This is an essential feature that makes the generated images highly applicable in oral diagnostics and dental education. Further validating the success of our approach, an evaluation panel composed of five experienced clinicians conducted an analysis of 30 $x_{female}^{syn}$ and 30 $x_{male}^{syn}$ images. The panel noted that the images produced by our application effectively achieved realistic de-identification, while simultaneously maintaining a high degree of realism in the preserved dental features. Furthermore, they evaluated these images as being of moderate usefulness for dental diagnosis and educational purposes, underscoring the practical value of our method in real-world applications.

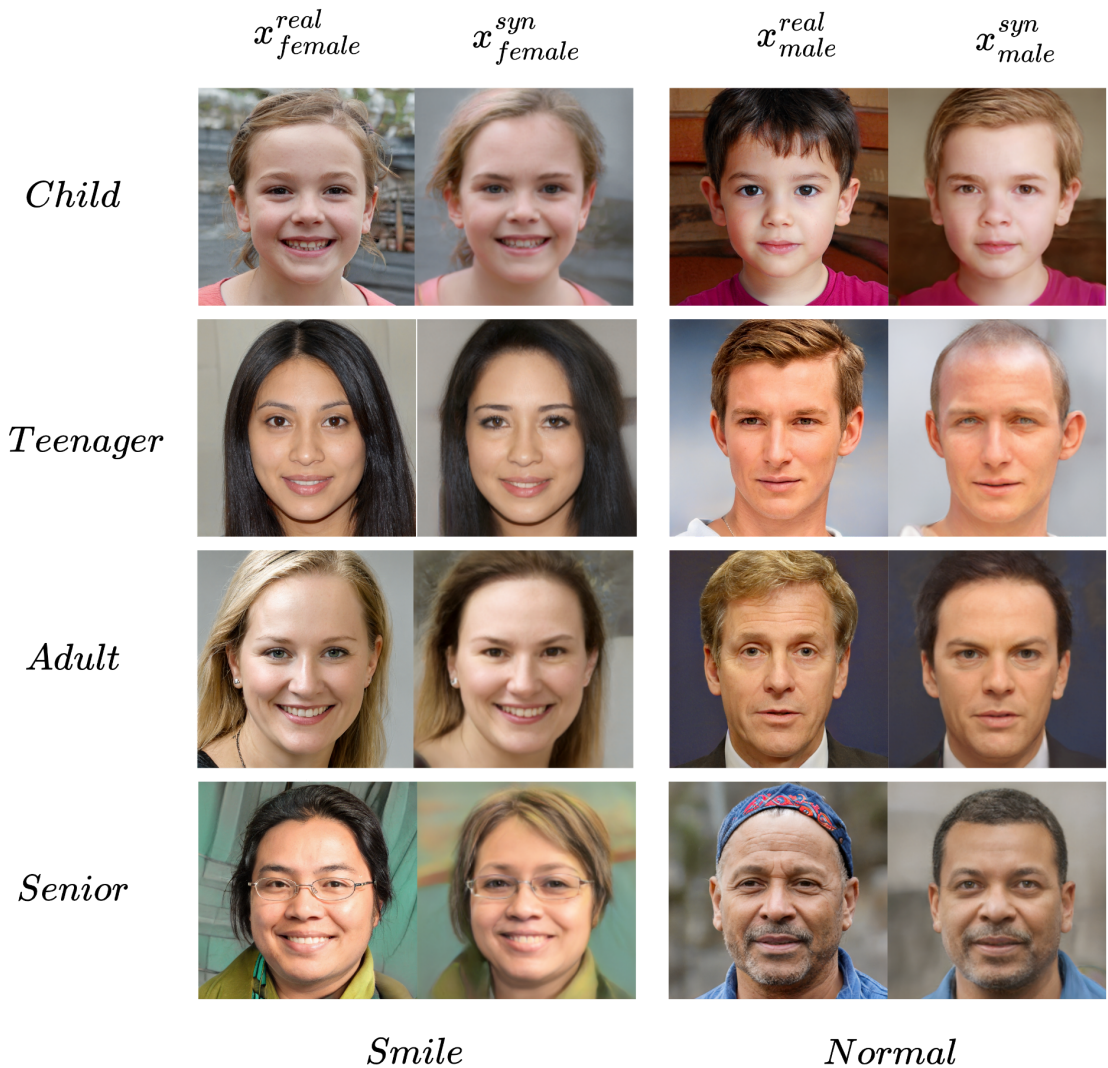

Fig.4. Results of our Area-preserving GAN Inversion across different gender and age groups

## Discussion

Although our study demonstrates the effective use of GAN inversion in safeguarding patient privacy while preserving dental features in images, a few limitations merit discussion. The first limitation is linked to our GAN inversion technique, which may not capture all subtle but essential dental features in each instance. This potential oversight could affect diagnostic accuracy and educational utility, making it a prime area for future refinement. The second limitation stems from the reliance on a limited Context Image Database of 300 manually labeled fake identities, which may constrain the diversity and realism of generated images. Expanding this database to encompass a wider range of identities could further improve the quality of synthesized images. Finally, while the identity matching process in our study effectively uses gender, age, and race for realistic image synthesis, it overlooks other important facial characteristics such as unique features and expressions. The inclusion of other distinguishing facial characteristics in future iterations could further enhance the realism and quality of the synthesized, de-identified images. Despite these limitations, our study presents a significant stepping stone, paving the way for future developments in privacy-preserving medical imaging.

## Conclusion

This study introduced the novel concept of an Area-preserving Encoder, designed to safeguard dental areas while modifying the facial identities in dental patient images. The foundation of this model is rooted in the state-of-the-art GAN-Inversion technique, recognized for its significant contributions to visual synthesis and image editing tasks. We also presented a comprehensive framework that utilizes deep learning models for critical tasks such as face localization, classification, and identity generation. This framework serves as an end-to-end guide, providing valuable insight into the development process of similar applications. This pipeline can be leveraged for future research and development initiatives, leading to innovative solutions in the realm of healthcare informatics. Despite a few limitations, this research provides the groundwork for a potential revolution in the handling of patient images, showcasing the power of artificial intelligence in promoting privacy while simultaneously supporting diagnostic and educational efforts in the field of dental healthcare.